\documentclass[conference]{IEEEtran}
\usepackage{cite}
\ifCLASSINFOpdf
  \usepackage[pdftex]{graphicx}
\else
\fi
\usepackage{amsmath}
\usepackage{array}
\ifCLASSOPTIONcompsoc
  \usepackage[caption=false,font=normalsize,labelfont=sf,textfont=sf]{subfig}
\else
  \usepackage[caption=false,font=footnotesize]{subfig}
\fi
\usepackage{fixltx2e}

\usepackage{stfloats}
\graphicspath{ {./images/} }
\begin{document}
\title{Gaussian Process Kernels for Popular State-Space Time Series Models}

\author{\IEEEauthorblockN{Alexander Grigorievskiy}
\IEEEauthorblockA{Department of Computer Science\\
Aalto University\\
Konemiehentie 2, Espoo, Finland \\
E-mail: alexander.grigorevskiy@aalto.fi}
\and
\IEEEauthorblockN{Juha Karhunen}
\IEEEauthorblockA{Department of Computer Science\\
Aalto University\\
Konemiehentie 2, Espoo, Finland \\
E-mail: juha.karhunen@aalto.fi}}

\maketitle

\begin{abstract}
In this paper we investigate a link between state-space models
and Gaussian Processes (GP) for time series modeling and forecasting. In particular,
several widely used state-space models are transformed into continuous
time form and corresponding Gaussian Process kernels are derived. Experimental
results demonstrate that the derived GP kernels are correct and appropriate for
Gaussian Process Regression. An experiment with a real world dataset shows that
the modeling is identical with state-space models and with the proposed GP kernels. 
The considered connection allows the researchers to look at their models from a different
angle and facilitate sharing ideas between these two different modeling approaches.
\end{abstract}

\IEEEpeerreviewmaketitle

\section{Introduction and Motivation}
Time series modeling and prediction is one of oldest topics in statistics. The very first statisticians already dealt with time dependent data. For example, Beveridge wheat price (years 1500 to 1869) or Wolfer's sunspot number (years 1610-1960)~\cite{yaglom_book1987} are examples of very early time series. Nowadays time series analysis and forecasting is ubiquitous in many fields of science and engineering. Econometricians, physicists, statisticians, biologists, climatologists etc. encounter time dependent data in their daily work.

Since this problem is very old and very wide-spread, different fields of science developed their own sets of methods for analysis and forecasting of time series. For instance, in statistics and econometrics domains the most common models are state-space (SS) models~\cite{durbin_book},~\cite{harvey_book}. In the physics domain the dominating class of models constitute nonlinear dynamical models~\cite{kantz_book_2004}. In the machine learning area time series are usually modeled by neural networks, fuzzy systems and Gaussian Processes. An overview of time series forecasting can be found in~\cite{gooijer_2006}.

One historically important subclass of the state-space models is autoregressive integrated moving average (ARIMA). It is still widely used and considered one of the best~\cite{box_book_2008} in time series analysis. A \textit{structural time series model} (STM) is a version of ARIMA where some time series components like trends and periodicities are imposed explicitly. It has an advantage over the pure ARIMA methodology that model misspecification is much less probable~\cite{harvey_book}. Moreover, STM is a way to introduce prior information and desired behavior into a time series model. Often a practitioner finds it difficult to consider and comprehend different forecasting methods from different domains. This paper is intended to shorten the gap between widely used STM models and Gaussian Processes (GPs) used in machine learning. The term structural time series model and state-space time series model are used interchangeably in this paper~\cite{durbin_book}. 

Basic state-space models are usually presented in the books~\cite{durbin_book},~\cite{harvey_book} as discrete time models with Gaussian errors. A \textit{structural time series} framework allows to combine several basic state-space models into more complex ones. There are generalizations of discrete-time SS models to continuous time~\cite[Chap. 9]{harvey_book} which after a certain procedure may be converted back to the discrete time. Since the errors in the basic SS models are assumed to be Gaussian, those are also GP models, however a direct systematic connection to Gaussian Processes used in machine learning is unknown to authors. The goal of this paper is to provide explicit connections between GP models and structural time series models.

Gaussian Processes are an important class of models in machine learning~\cite{rasmussen_2005}. Modeling of time series has been widely addressed by GP community~\cite{roberts_2012}. The modeling principles differ significantly from the state-space models. Modeling is done in continuous time and the main object to model is covariance function (and optionally mean function). There exist a known connection between continuous-discrete state space model and Gaussian process~\cite{hartikainen2010}. The advantage of representing the GP in SS form is that the inference can be done in $O(N)$ time where $N$ is the number of data points, while the classic GP regression requires $O(N^3)$ operations. However, if the amount of data points is relatively small $N <10000$, or we use some modification of standard GP, the difference in computational time can become negligible~\cite{ambikasaran_2014} on modern computers.

In this paper we derive several GP covariance functions which correspond to the main structural time series models. This explicit connection is useful for the researches with different background. State-space modelers can see that their methods are equivalent to certain Gaussian Processes and they can try to use various extension developed in the GP literature. GP specialists on the other hand can analyze the covariance functions corresponding to state-space models and borrow some ideas from there.

\section{Structural Time Series Models and Gaussian Processes}\label{sec:gp}
Random (Stochastic) process is a collection of random variables $X_t, t \in T$ parametrized by the set $T$. If $T$ is a set of integers ($T = \mathcal{Z}$) then the random process is discrete. If is real-valued ($T = \mathcal{R}$) the process is continuous. 

The random process can be completely described by the infinite number of distribution function of the form $F_N(v_1, v_2, \cdots, v_N) = {\mathtt{Pr}[X(t_1)< v_1, X(t_2)< v_2, \cdots, X(t_n)< v_N]}$ for any positive integer $N$ and arbitrary selected time points $t_1, t_2, \cdots, t_N$. Although this description is complete it is cumbersome. Therefore, often in practice only the first two distribution functions are taken into account. 

These first two distribution functions allow to define the first moments of the random process: mean and covariance. Using these first two moments we can define the important class of random processes - Wide-Sense Stationary (WSS) Random Process. For a random process to be WSS it is sufficient that the mean is constant, variance is finite, and covariance function depends only on difference between time points. More detailed information can be found in any book about stochastic processes e.g.~\cite{yaglom_book1987}.

\subsection{Gaussian Process (GP)}
A \textit{Gaussian process} is a random process $f(t)$ where for arbitrary selected time points $t_1, t_2,\cdots, t_N$ the probability distribution $\mathtt{p}[f(t_1), f(t_2), \cdots f(t_N)]$ is multivariate Gaussian.

To define a Gaussian process it is necessary to define a mean function $m(t) = \mathtt{E}[f(t)]$ and covariance function $\mathtt{Cov[t_1, t_2]} = \mathtt{E}[(f(t_1) - m(t_1))(f(t_2) - m(t_2))]$.

\subsection{State-Space Models}
The state-space model is the model of the form:
\begin{equation}\label{eq:1}
\begin{aligned}
&\mathbf{z}_n = A_{n-1} \mathbf{z}_{n-1} + \mathbf{q}_n \quad \text{(state / dynamic equation)} \\
&y_n = H_n \mathbf{z_n} + \epsilon_n \quad \text{(measurement equation)}
\end{aligned}
\end{equation}
It is assumed that $y_n$ (scalar) are the observed values of this random process. The noise terms $\mathbf{q}_n$ and $\epsilon_n$ are, in basic case, assumed to be Gaussian. This is the assumption we do in this paper. When the noise terms are Gaussian the random process $y_n$ is also Gaussian and we find the explicit form of covariance function for the most popular state-space models.

The Kalman filter algorithm allows to make inference about the model~(\ref{eq:1}). It computes the different conditional distributions of the hidden state $\mathbf{z_n}$ as well as a likelihood of the model~\cite{sarkka_book}.

In the model~(\ref{eq:1}) the state variable $\mathbf{z}_n$ is assumed to be discrete. There exist equivalent versions where the state variable is continuous and it is called continuous-discrete state-space model~\cite{harvey_book}. The relationships between continuous-discrete state-space models and Gaussian processes have been recently highlighted~\cite{hartikainen2010}. In this paper the connection is made more explicit and clear.

\subsection{Combining Models / Structural Time-Series (STS) Models }\label{sec:combine}

The structural time series framework is a way to construct state-space models and incorporate the desired properties or prior information into them. These properties are fixed level, trend, periodicity and quasi-periodicity (cyclicity)~\cite{durbin_book},~\cite{harvey_book}. The ability to incorporate prior information is an advantage of the STS modeling framework over more general ARIMA approach. A certain state-space model corresponds to each aforementioned property. Let's show how to combine these models additively. Suppose that $y_n = z_n^{trend} + z_n^{periodic} + \epsilon_n$, so $y_n$ is a sum of trend and periodic component. It is possible to write it in a single state-space model: 

\begin{equation}\label{eq:2}
\begin{aligned}
& \begin{bmatrix} \mathbf{z}_n^{(tr)} \\ \mathbf{z}_n^{(per)} \end{bmatrix} = 
\begin{bmatrix} A_{n-1}^{(tr)} & 0 \\ 0 & A_{n-1}^{(per)} \end{bmatrix} 
\begin{bmatrix} \mathbf{z}_{n-1}^{(tr)} \\ \mathbf{z}_{n-1}^{(per)} \end{bmatrix}+ 
\begin{bmatrix} \mathbf{q}_n^{(tr)} \\ \mathbf{q}_n^{(per)} \end{bmatrix}\\
&y_n = [H_n^{(tr)} H_n^{(per)}] \begin{bmatrix} \mathbf{z}_{n-1}^{(tr)} \\ \mathbf{z}_{n-1}^{(per)} \end{bmatrix} + \epsilon_n
\end{aligned}
\end{equation}

It can be easily seen that $\mathbf{z}_n^{(tr)}$ and $\mathbf{z}_n^{(per)}$ are uncorrelated random processes if their noise terms are uncorrelated. In this case the covariance function of $y_n$ is:

\begin{equation}\label{eq:3}
\begin{aligned}
& \mathtt{Cov}[y_k, y_{k+n}] = H_n^{(tr)} \mathtt{Cov}[z_k^{(tr)}, z_{k+n}^{(tr)}] (H_n^{(tr)})^T + \\
& + H_n^{(per)} \mathtt{Cov}[z_k^{(per)}, z_{k+n}^{(per)}] (H_n^{(per)})^T + \delta_{(n=0)} \sigma^2_{\epsilon}
\end{aligned}
\end{equation}

Here $\delta_{(n=0)}$ is a Kronecker delta which equals 1 when $n=0$.
So, the covariance is a sum of two covariances (matrices $H$ are often 1) and a white noise term from the measurement equation. 
This useful property will be utilized in the subsequent sections.

\section{Bayesian Linear Regression in State-Space Form}\label{sec:blr}
At first, recall the Bayesian Linear Regression (BLR) in the state-space form. Assume that we have $N$ measurements $\mathbf{y} = [y_1, y_2, \hdots, y_N]^T$, which are observed at time points $\mathbf{t} = [t_1, t_2, \hdots, t_N]^T$. Further, assume that there is a linear dependency between measurements and time: 

\begin{equation}\label{eq:4}
\begin{split}
&y_k = \theta t_k + \epsilon_k\\
&\theta \sim \mathcal{N}(m_0, P_0) \quad \text{- prior of the parameter $\theta$}\\
&\epsilon_k \sim \mathcal{N}(0, \sigma^{2}_0) \quad \text{- Gaussian white noise}\\
\end{split}
\end{equation}.

$\theta$ is a parameter of the model and the prior for it is $\theta \sim \mathcal{N}(m_0, P_0)$, $\epsilon$ is a Gaussian white noise: $\epsilon \sim \mathcal{N}(0, \sigma^{2}_0)$. In this formulation, the BLR provides us the posterior distribution of $\theta$ which we are not currently interested in. Besides, it provides the posterior predictive distribution which for any set of time points $t^{\star}_1, t^{\star}_2, \hdots, t^{\star}_M$ yields the distribution of corresponding measurements. It is well know\cite{rasmussen_2005} that the same posterior predictive distribution can be obtained by Gaussian Process Regression (GPR) with the kernel:

\begin{equation}\label{eq:5}
\mathbf{y} \sim 
\mathcal{GP} \left( m_0 \mathbf{t}, P_0 \mathbf{t} \mathbf{t}^T + \sigma_0^2 I \right)
\end{equation}

We are interested in representing the BLR model in the state-space form because it allows us to look at the model in the sequential form when data arrives one by one. Moreover, the Kalman filter type inference which is the standard for the linear state-space models scales linearly with the number of samples, while Gaussian Process or batch BLR scales cubically\cite{rasmussen_2005}. There are several ways to express BLR in the state-space form, the one we are interested in is written below~\cite[p. 37]{sarkka_book}:

\begin{equation}\label{eq:6}
\begin{aligned}
&\begin{cases}
&\begin{bmatrix} x_k \\ \theta_k \end{bmatrix} = \begin{bmatrix}  1 & \Delta t_{k-1} \\ 0 & 1 \end{bmatrix}  
\begin{bmatrix} x_{k-1} \\ \theta_{k-1} \end{bmatrix}  \\
&y_k = \begin{bmatrix} 1 & 0 \end{bmatrix} \begin{bmatrix} x_k \\ \theta_k \end{bmatrix} + \epsilon_k, \quad \text{where:}
\end{cases}\\
&x_0 = 0 \sim \mathcal{N}(0,0), \quad \theta_0 \sim \mathcal{N}(m_0, P_0), \quad \epsilon_k \sim \mathcal{N}(0, \sigma^{2}_0) \\
&\Delta t_{k-1} = t_k - t_{k-1}, \quad \text{and it is assumed that $t_0 = 0$.}
\end{aligned}
\end{equation}

Now let's check that the state-space model listed above is indeed equivalent to Bayesian Linear Regression. Looking at the equation for $\theta$ we see that $\theta_{k} = \theta_{k-1}$ for all $k$, so it does not change with time. Since $t_0 = 0 \text{ and } x_0 = 0$ we have that:
\begin{align*}
x_1 &= t_1 \theta_0 = \theta t_1 \\
x_2 &= x_1 + (t_2 - t_1) \theta_1 = t_2 \theta_1 + t_1(\theta_0-\theta_1) = t_2 \theta_1 = \theta t_2 \\
\vdots \\ 
x_k &= x_{k-1} + (t_k - t_{k-1}) \theta_{k-1} \\
&= t_{k} \theta_{k-1} + t_{k-1}(\theta_{k-2}-\theta_{k-1}) = t_{k} \theta_{k-1} = \theta t_{k} \\
\end{align*}

So, we see that $x_k = t_k \theta$ and if we insert the obtained result into the equation for $y_k$: $y_k = \theta t_k + \epsilon_k$
which exactly coincides with the original BLR formulation. Using the obtained state-space model we can find the covariance matrix of $y_k$. It would be the same as the one in Eq.~(\ref{eq:5}). We are going to explicitly derive the covariance function for the more general state-space model in the next section.

In this section we have shown the equivalence of Gaussian Process Regression with covariance matrix in Eq.~(\ref{eq:5}) and state-space formulation in Eq.~(\ref{eq:6}). These two models are also equivalent to the Bayesian Linear Regression.

\section{General State-Space Model with Random Noise}\label{sec:4}
In this section we derive the covariance function form for a more general state-space model than in the previous section. In the literature this model is called \textit{Local Linear Trend Model} (LLLM). It is shown that this general state-space model under the special setting of parameters becomes equivalent to the well-known time series models: local level model, BLR, connection with the quasi-periodic (cyclic) model is very close as well. Derivation of covariance function provides us a useful connection to the Gaussian Process Regression for the aforementioned models. The general state-space model is:  

\begin{equation}\label{eq:7} 
\begin{aligned}
&\begin{cases}
&\begin{bmatrix} x_k \\ \theta_k \end{bmatrix} = \begin{bmatrix}  1 & \Delta t_{k-1} \\ 0 & 1 \end{bmatrix}  
\begin{bmatrix} x_{k-1} \\ \theta_{k-1} \end{bmatrix} + \begin{bmatrix} q_k^{(1)} \\ q_k^{(2)} \end{bmatrix}\\
&y_k = \begin{bmatrix} 1 & 0 \end{bmatrix} \begin{bmatrix} x_k \\ \theta_k \end{bmatrix} + \epsilon_k, \quad \text{where:} \quad \epsilon_k \sim \mathcal{N}(0, \sigma^{2}_0)\\
\end{cases}\\
&\Delta t_{k-1} = t_k - t_{k-1}, \quad \text{it is assumed that $t_0 = 0$,} \\
&\begin{bmatrix} x_0 \\ \theta_0 \end{bmatrix} \sim \mathcal{N}\left( \begin{bmatrix} c_0 \\ m_0 \end{bmatrix},
 \begin{bmatrix} K_0 & 0 \\ 0 & P_0 \end{bmatrix} \right) \\
&\begin{bmatrix} q_k^{(1)} \\ q_k^{(2)} \end{bmatrix} \sim \mathcal{N}\left( \begin{bmatrix} 0 \\ 0 \end{bmatrix},
 \begin{bmatrix} q_0^2 \Delta t_{k-1} & 0 \\ 0 & g_0^2 \Delta t_{k-1} \end{bmatrix} \right) \\
\end{aligned}
\end{equation}

As we can see the difference with the state-space model from the previous section consist of extra noise terms in the dynamic (or state) equation. Another difference is non-zero prior distribution for the initial state variable $x_0$. Now it is distributed as a Gaussian random variable: $x_0 \sim \mathcal{N}(c_0, K_0)$.

\subsection{Noise in Dynamic Equation}\label{sec:4_1}
In this subsection the extra noise terms which appear in the dynamic equation are briefly discussed. In the two dimensional noise term $\mathbf{q} = \begin{bmatrix} q_k^{(1)} \\ q_k^{(2)} \end{bmatrix}$ the two components are independent and Gaussian distributed. Consider, for example, the first component $q_k^{(1)} \sim \mathcal{N}(0, q_0^2 \Delta t_{k-1})$. It is a classical \textit{Wiener process}\cite{rasmussen_2005} also called \textit{standard Brownian motion} and is a generalization of a simple random walk to the continuous time when time measurements are not necessary equidistant. Its covariance function is $\mathtt{Cov}[q_k^{(1)}(t_1), q_k^{(1)}(t_2)] = K_0 + q_0^2 \min(t_1,t_2)$ and it is a basic example of nonstationary Gaussian Process.
 
\subsection{Covariance Function Derivation}
Before commencing the derivation of the covariance function we consider an important property of the state-space model in Eq.~(\ref{eq:7}). Denote:

\begin{equation} \label{eq:8}
A[\Delta t_{k-1}] = \begin{bmatrix}  1 & \Delta t_{k-1} \\ 0 & 1 \end{bmatrix}
\end{equation}

We can easily verify that:
\begin{equation}\label{eq:9}
\begin{aligned}
A[\Delta t_{k}] & A[\Delta t_{k-1}] = \begin{bmatrix}  1 & \Delta t_{k} \\ 0 & 1 \end{bmatrix} \begin{bmatrix}  1 & \Delta t_{k-1} \\ 0 & 1 \end{bmatrix} =\\ 
&= \begin{bmatrix}  1 & \Delta t_{k} + \Delta t_{k-1} \\ 0 & 1 \end{bmatrix} =  A[\Delta t_{k} + \Delta t_{k-1}] 
\end{aligned} 
\end{equation}

This is a convenient property which will be utilized during the derivation and later in the Sec.~\ref{sec:5}.
Consider the covariance function: 

\begin{equation}\label{eq:10}
\begin{aligned}
&\mathtt{Cov}[y_k, y_{k+n}] = \mathtt{E}[ (y_k - \mathtt{E}[y_k]) (y_{k+n} - \mathtt{E}[y_{k+n}])] = \\
&= \mathtt{E}[ (x_k + \epsilon_k - \mathtt{E}[x_k]) (x_{k+n} + \epsilon_{k+n} - \mathtt{E}[x_{k+n}])] = \\ 
&= \mathtt{Cov}[x_k, x_{k+n}] + \delta_{(n=0)} \sigma_0^2 
\end{aligned}
\end{equation}

 Therefore, we see that in order to find the covariance function of $y_k$ it is enough to find the covariance function of $x_k$ and add the Kronecker symbol mentioned above. So, we can ignore the measurement equations right now and write our state-space model in the vector form:

\begin{equation}
\mathbf{z}_k = A[\Delta t_{k-1}] \mathbf{z}_{k-1} + \mathbf{q}_k;  \;\;  \mathbf{z}_k = \begin{bmatrix} x_k \\ \theta_k \end{bmatrix}; \;\; \mathbf{q}_k = \begin{bmatrix} q_k^{(1)} \\ q_k^{(2)} \end{bmatrix} 
\end{equation}.

Lets express $z_k$ through the initial conditions and noise terms:

\begin{equation}\label{eq:11}
\begin{aligned}
&\mathbf{z}_k = A[\Delta t_{k-1}] \mathbf{z}_{k-1} + \mathbf{q}_k = \\
&= A[\Delta t_{k-1}] ( A[\Delta t_{k-2}] \mathbf{z}_{k-2} + \mathbf{q}_{k-1}) + \mathbf{q}_k = \\
&= A[\Delta t_{k-1} + \Delta t_{k-2}] \mathbf{z}_{k-2} + A[\Delta t_{k-1}] \mathbf{q}_{k-1} + \mathbf{q}_k = \hdots = \\
&= A[\Delta t_{k-1} + \Delta t_{k-2} + \hdots + \Delta t_{0} ] \mathbf{z}_{0} + A[\Delta t_{k-1} + \hdots \\
&\hdots + \Delta t_{k-2} + \Delta t_{1} ] \mathbf{q}_{1} + \hdots + A[\Delta t_{k-1}] \mathbf{q}_{k-1} + \mathbf{q}_{k}
\end{aligned}
\end{equation}

Here we use the property from Eq.~(\ref{eq:9}) of the transition matrix. We see that $\mathbf{z}_k$ is a sum of terms each of which is a vector times matrix $A$ with different arguments. Vectors are $\mathbf{z}_{0}, \mathbf{q}_{0} \cdots \mathbf{q}_{k}$. Arguments of matrix $A$ are also sum of terms $\Delta t_{i}$ and the number of terms decreases one by one from $k$ in $A[\Delta t_{k-1} + \Delta t_{k-2} + \hdots + \Delta t_{0} ]$ to zero in front of $\mathbf{q}_k$. We can easily compute the mean of $\mathbf{z}_k$, taking into account the fact that the mean $\mathtt{E}[\mathbf{q}_i] = 0$ and expanding the expressions for $\Delta t_{i}$

\begin{equation} \label{eq:13}
\mathtt{E}[\mathbf{z}_k] = A[\Delta t_{k-1} + \Delta t_{k-2} + \hdots + \Delta t_{0} ] \mathtt{E}[ \mathbf{z}_{0}] = 
A[ t_k ] \begin{bmatrix} c_0 \\ m_0 \end{bmatrix}
\end{equation}

Having the expression for $\mathbf{z}_k$ and its mean we can compute the covariance $\mathtt{Cov}[\mathbf{z}_k, \mathbf{z}_{k+n}] = \mathtt{E}[ (\mathbf{z}_k - \mathtt{E}[\mathbf{z}_k]) (\mathbf{z}_{k+n}^T - \mathtt{E}[\mathbf{z}_{k+n}^T])]$. The computation is quite straightforward using Eq.~(\ref{eq:11}) and the fact that $\mathbf{z}_{0}$ and all $\mathbf{q}_{i}$ are mutually independent. The final answer is presented below:

\begin{equation}  \label{eq:14}
\begin{aligned}
&\mathtt{Cov}[\mathbf{z}_k, \mathbf{z}_{k+n}] = A[\Delta t_{k-1} + \Delta t_{k-2} + \hdots + \Delta t_{0} ] \, \mathtt{Cov}[\mathbf{z}_0, \mathbf{z}_{0}] \, \\ 
&A[\Delta t_{k+n-1} + \Delta t_{k+n-2} + \hdots + \Delta t_{0} ]^T + A[\Delta t_{k-1} + \hdots \\
&+\Delta t_{k-2}+\Delta t_{1} ] \, \mathtt{Cov}[\mathbf{q}_1, \mathbf{q}_{1}] \, A[\Delta t_{k+n-1} + \Delta t_{k+n-2} + \hdots \\
&+\Delta t_{1} ]^T + \cdots + \mathtt{I} \, \mathtt{Cov}[\mathbf{q}_k, \mathbf{q}_{k}] \, A[\Delta t_{k+n-1} + \hdots + \Delta t_{k} ]^T 
\end{aligned}
\end{equation}

As we see the expression is the sum of terms $ A[\cdot] \, \mathtt{Cov}[\cdot,\cdot] \, A[\cdot]^T$ where the arguments of $A[\cdot]$ and $A[\cdot]^T$ are different while arguments in $\mathtt{Cov}[\cdot,\cdot]$ are the same.

Now suppose we want to compute all possible covariances up to some maximal time index $N$, i. e. $\mathtt{Cov}[\mathbf{z}_k, \mathbf{z}_{n}], \; \text{where} \; 1 \leq k \leq N, \; 1 \leq n \leq N$. These covariances can be written in a matrix consisting of $2 \times 2$ blocks (because $\mathtt{Cov}[\mathbf{z}_k, \mathbf{z}_{n}]$ - one block is $2 \times 2$), and so in total it is a $2N \times 2N$ matrix. In the next formula we present the form of this matrix, and later the expression for the single components are provided. To simplify the notations and make them more vivid, suppose $N = 3$: 

\begin{equation} \label{eq:15}
\begin{aligned} 
& \begin{bmatrix}
\mathtt{Cov}[\mathbf{z}_0, \mathbf{z}_{0}] & \mathtt{Cov}[\mathbf{z}_0, \mathbf{z}_{1}] & \mathtt{Cov}[\mathbf{z}_0, \mathbf{z}_{2}] & \mathtt{Cov}[\mathbf{z}_0, \mathbf{z}_{3}] \\   
\mathtt{Cov}[\mathbf{z}_1, \mathbf{z}_{0}] & \mathtt{Cov}[\mathbf{z}_1, \mathbf{z}_{1}] & \mathtt{Cov}[\mathbf{z}_1, \mathbf{z}_{2}] & \mathtt{Cov}[\mathbf{z}_1, \mathbf{z}_{3}] \\ 
\mathtt{Cov}[\mathbf{z}_2, \mathbf{z}_{0}] & \mathtt{Cov}[\mathbf{z}_2, \mathbf{z}_{1}] & \mathtt{Cov}[\mathbf{z}_2, \mathbf{z}_{2}] & \mathtt{Cov}[\mathbf{z}_2, \mathbf{z}_{3}] \\
\mathtt{Cov}[\mathbf{z}_3, \mathbf{z}_{0}] & \mathtt{Cov}[\mathbf{z}_3, \mathbf{z}_{1}] & \mathtt{Cov}[\mathbf{z}_3, \mathbf{z}_{2}] & \mathtt{Cov}[\mathbf{z}_3, \mathbf{z}_{3}]
\end{bmatrix} =\\
& \qquad \qquad \qquad = \mathcal{P}\{T\} \; D_0 \; (\mathcal{P}\{T\})^T \\
\end{aligned}
\end{equation}

We are not interested in computing the first row and column of this covariance matrix since variable $\mathbf{z}_0$ does not correspond to any real observation, it is just an initial random variable. Also, $\mathcal{P}\{T\}$ in the above formula equals:
 
\begin{equation}
\begin{aligned} 
&\mathcal{P}\{T\} = \\ 
&\begin{bmatrix}
A[0] & 0 & 0 & 0\\
A[\Delta t_{0}] & A[0] & 0 & 0\\

A[\Delta t_{1} + \Delta t_{0}] & A[\Delta t_{1}] & A[0] & 0 \\

A[\Delta t_{2} + \Delta t_{1} + \Delta t_{0}] & A[\Delta t_{2} + \Delta t_{1}] & A[\Delta t_{2}] & A[0]\\
  
\end{bmatrix}
\end{aligned}
\end{equation}

Here each element of the $\mathcal{P}\{T\}$ matrix is a $(2 \times 2)$ block. The notation $\mathcal{P}\{T\}$ means that some matrix operator $\mathcal{P}\{ \cdot \}$ is applied to the matrix $T$. Currently we are not specifying
what are $\mathcal{P}\{\cdot\}$ and $T$, it is done later in this section when we obtain covariances of $x_k$.

Matrix $D_0$ in Eq.~(\ref{eq:15}) is a block diagonal matrix written below:
\begin{equation} 
D_0 = 
\begin{bmatrix}
\mathtt{Cov}[\mathbf{z}_0, \mathbf{z}_{0}] & 0 & 0 & 0\\
0 & \mathtt{Cov}[\mathbf{q}_1, \mathbf{q}_{1}] & 0 & 0\\
0 & 0 & \mathtt{Cov}[\mathbf{q}_2, \mathbf{q}_{2}] & 0\\
0 & 0 & 0 & \mathtt{Cov}[\mathbf{q}_3, \mathbf{q}_{3}] \\
\end{bmatrix}
\end{equation}

It can be verified that expressions in Eq.~(\ref{eq:14}) and Eq.~(\ref{eq:15}) are equal. Covariances $\mathtt{Cov}[\mathbf{q}_i, \mathbf{q}_{i}]$ are diagonal matrices shown in Eq.~(\ref{eq:7}). Thus, we have derived the expression for $\mathtt{Cov}[\mathbf{z}_k, \mathbf{z}_{n}]$. However we are not interested in it as is. We would like to know the covariances $\mathtt{Cov}[x_k, x_{n}]$ because they are directly related with covariances of the observed variable $y_k$ which is shown in Eq.~(\ref{eq:10}). It means that we are interested in the covariance matrix consisting of odd columns and rows of the matrix $\mathcal{P}\{T\} \; D_0 \; (\mathcal{P}\{T\})^T$. To derive it consider the structure of the expression which is the main building block in covariance functions in Eq.~(\ref{eq:14}) and Eq.~(\ref{eq:15}):

\begin{equation}\label{eq:16}
\begin{aligned}
&A \left[ \sum \Delta t_{m} \right] \begin{bmatrix}  q^{2}_0 \Delta t_{i-1} & 0 \\ 0 & g^{2}_0 \Delta t_{i-1} \end{bmatrix} A^{T} \left[ \sum \Delta t_{n} \right] = \\
&=\begin{bmatrix}  1 & \sum \Delta t_{m} \\ 0 & 1 \end{bmatrix} \begin{bmatrix}  q^{2}_0 \Delta t_{i-1} & 0 \\ 0 & g^{2}_0 \Delta t_{i-1} \end{bmatrix} \begin{bmatrix}  1 & 0 \\ \sum \Delta t_{n} & 1 \end{bmatrix} = \\
&= \begin{bmatrix}  \boxed{ q^{2}_0 \Delta t_{i-1} +  g^{2}_0 \Delta t_{i-1} \left( \sum \Delta t_{m} \right) \left( \sum \Delta t_{n} \right)} & \cdot \\ \cdot & \cdot \end{bmatrix}
\end{aligned}
\end{equation}

In the above formula we are interested in the top left element which is emphasized by the rectangle, because it gives exact covariances of $\mathtt{Cov}[x_k, x_{n}]$ from the covariances $\mathtt{Cov}[\mathbf{z}_k, \mathbf{z}_{n}]$. Now we see that the required covariance consist of two parts which correspond to the two terms in the sum above. The first term is affected only by the top diagonal entry of the middle matrix in the initial product. The second term is affected by the bottom diagonal entry and the arguments of the matrices $A$. Now we are ready to write the required correlations by looking at Eq.~(\ref{eq:14}), analyzing contributions of each term there and taking into account Eq.~(\ref{eq:16}). Representation~(\ref{eq:15}) is also useful in deriving the second part of the following result:

\begin{equation} \label{eq:17}
\begin{aligned}
&\begin{bmatrix}  
\mathtt{Cov}[x_1, x_1] & \mathtt{Cov}[x_1, x_2] & \mathtt{Cov}[x_1, x_2] \\ 
\mathtt{Cov}[x_2, x_1] & \mathtt{Cov}[x_2, x_2] & \mathtt{Cov}[x_2, x_3] \\
\mathtt{Cov}[x_3, x_1] & \mathtt{Cov}[x_3, x_2] & \mathtt{Cov}[x_3, x_3]
\end{bmatrix} 
 = \mathtt{Cov}_1[ \cdot ] + \mathtt{Cov}_2[ \cdot ]
\end{aligned}
\end{equation}

where:
\begin{equation}
\begin{aligned}
& \mathtt{Cov}_1[x_k, x_{k+n}] = K_0 + q_0^2 \Delta t_{0} + q_0^2 \Delta t_{1} + \hdots \\
& +  q_0^2 \Delta t_{k-1} = K_0 + q_0^2 t_k
\end{aligned}
\end{equation} 

Another way to write $\mathtt{Cov}_1[x_k, x_{k+n}]$ is:
\begin{equation}\label{eq:18}
\mathtt{Cov}_1[x_k, x_{k+n}] = K_0 + q_0^2 \min( x_k, x_{k+n} )
\end{equation}

The expression for $\mathtt{Cov}_2[ \cdot ]$ is written below:
\begin{equation} \label{eq:19}
\begin{aligned}
&\mathtt{Cov}_2[ \cdot ] = T D T^T \\
& \text{where:} \\ 
& T = \begin{bmatrix}
0 & 0 & 0 & 0\\
\Delta t_{0} & 0 & 0 & 0\\
\Delta t_{1} + \Delta t_{0} & \Delta t_{1} & 0 & 0\\
\Delta t_{2} + \Delta t_{1} + \Delta t_{0} & \Delta t_{2} + \Delta t_{1} & \Delta t_{2} & 0\\
\end{bmatrix} \\
& D = \begin{bmatrix}
P_0 & 0 & 0 & 0\\
0 & g_0^2 \Delta t_{0} & 0 & 0\\
0 & 0 & g_0^2 \Delta t_{1} & 0\\
0 & 0 & 0 & g_0^2 \Delta t_{2}\\
\end{bmatrix} 
\end{aligned}
\end{equation}

The matrix $T$ can also be represented as:

\begin{equation} 
T = 
\begin{bmatrix}
0 & 0 & 0 & 0\\
1 & 0 & 0 & 0\\
1 & 1 & 0 & 0\\
1 & 1 & 1 & 0\\  
\end{bmatrix}
\begin{bmatrix}
0 & 0 & 0 & 0\\
\Delta t_{0} & 0 & 0 & 0\\
\Delta t_{1}& \Delta t_{1} & 0 & 0\\
\Delta t_{2}& \Delta t_{2} & \Delta t_{2} & 0\\  
\end{bmatrix}
\end{equation}

\begin{figure}[h]
\centering
\subfloat[~\mbox{Brownian motion:} \hfill ${K_0 = 1, \underline{P_0=0},} {q^2_0=1, \underline{g^2_0=0}}$]{\includegraphics[width=0.44\columnwidth]{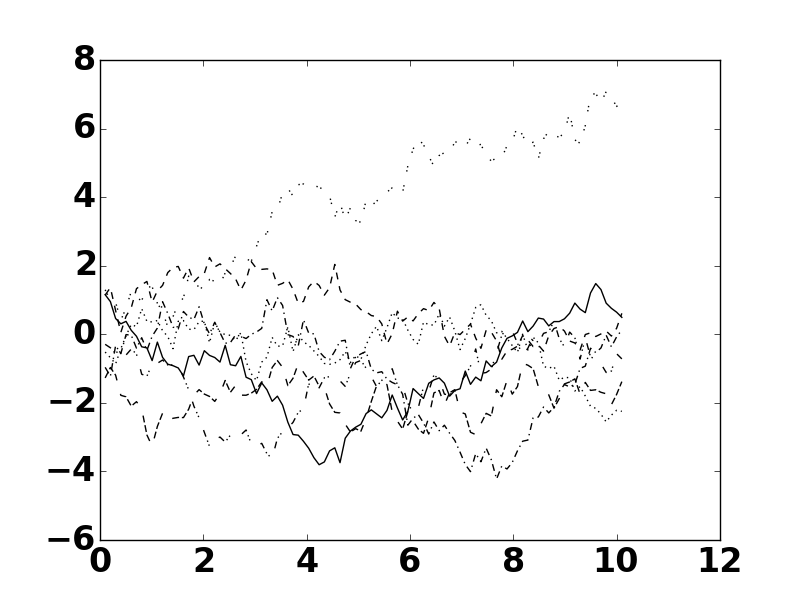}%
\label{fig_1a}}
\hfil
\subfloat[\mbox{Linear Regression:} \hfill ${K_0 = 1, P_0=1,} {\underline{q^2_0=0}, \underline{g^2_0=0}}$]{\includegraphics[width=0.44\columnwidth]{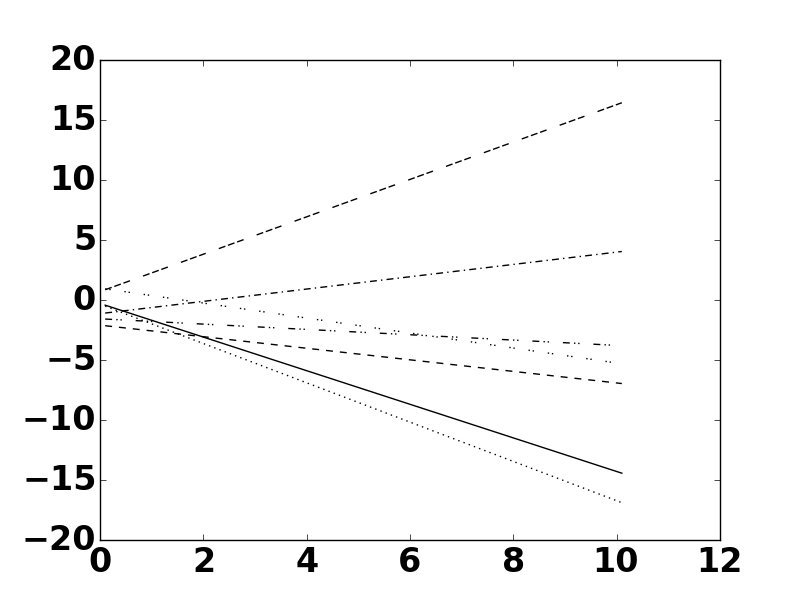}%
\label{fig_1b}}

\subfloat[${K_0 = 1, P_0=1,} {q^2_0=1, \underline{g^2_0=0}}$]{\includegraphics[width=0.44\columnwidth]{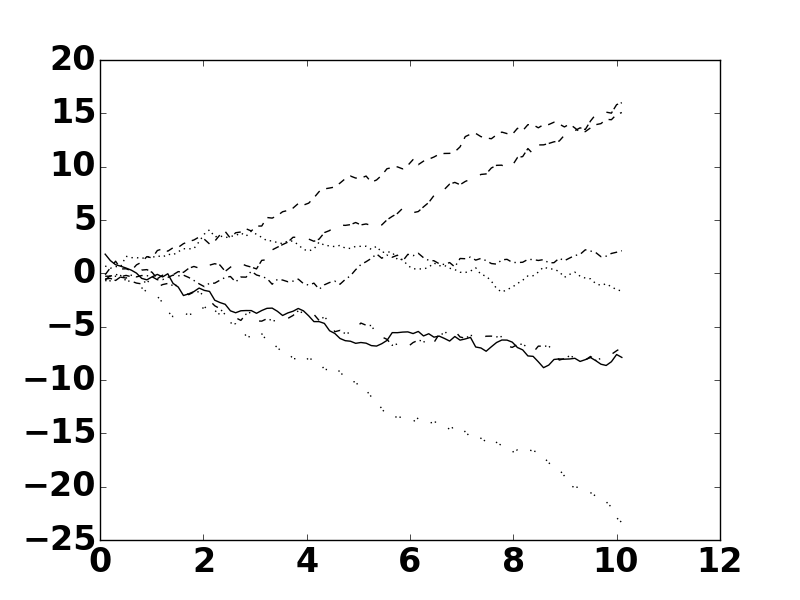}%
\label{fig_1c}}
\hfil
\subfloat[${K_0 = 1, P_0=1,} \; {q^2_0=1, g^2_0=1}$]{\includegraphics[width=0.44\columnwidth]{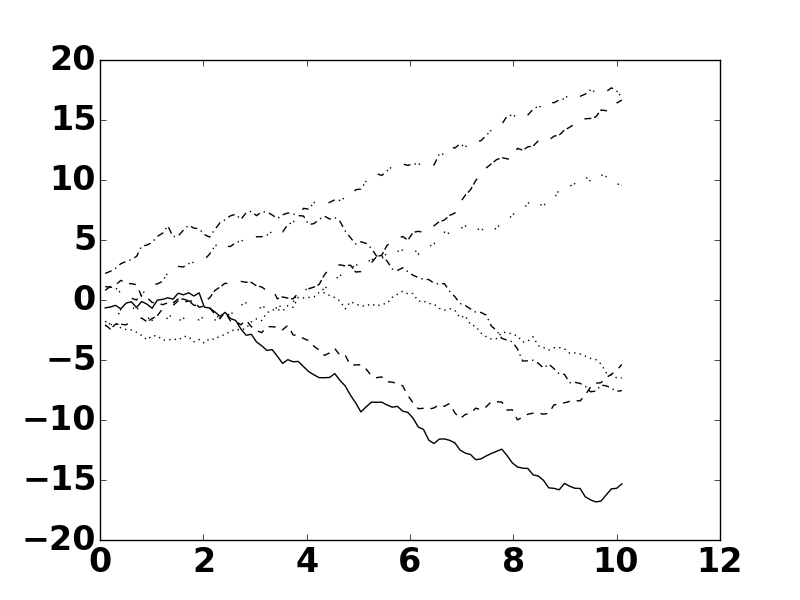}%
\label{fig_1d}}
\caption{GP sample paths of general covariance (LLLM) for various parameter values. Underline emphasize parameters which equal zero.}
\label{fig_gen_paths}
\end{figure}

In the Eq.~(\ref{eq:19}) we must ignore the first row and the first column so that the resulting $\mathtt{Cov}_2[ \cdot ]$ matrix is $3 \times 3$. It is possible to write this formula directly by $3 \times 3$ matrices but this form is useful for the derivation of quasi-periodic (cyclic) covariance in the next section. The Eq.~(\ref{eq:17}) is the final answer for the covariance function of the model stated in Eq.~(\ref{eq:7}). Now given time points $\mathbf{t} = [t_1, t_2, \hdots, t_N]^T$ we can compute the covariance function, the mean function which is given in Eq.~(\ref{eq:13}) and use Gaussian Process Regression in a regular way. The sample paths from GP with this covariance function are presented on Fig.~\ref{fig_gen_paths}.

Several standard structural time series models are actually versions of the general model described above, they are listed later in this section. The Bayesian Linear Regression model considered in Eq.~(\ref{eq:6}) in Section~\ref{sec:blr} is a lucid representative as well. If we set $c_0 = 0, K_0 = 0, q_0^2 = g_0^2 = 0$ as in the expression for BLR, then $\mathtt{Cov}_1 = 0$ and in $\mathtt{Cov}_2$ only the first element in the diagonal matrix is non-zero. Then, expanding all $\Delta t$ we easily get that the covariance becomes equal to the one of BLR (Eq.~\ref{eq:5}).

\subsection{Local Level Model}
Local Level Model (LLM) is the simplest model among the structural time series models~\cite{durbin_book}. Its standard representation in the literature is:
\begin{equation}\label{eq:20} 
\begin{aligned}
&\begin{cases}
&x_k = x_{k-1} + q_k; \qquad q_k \sim \mathcal{N}(0, q_0^2) \\
&y_k = x_k + \epsilon_k;\qquad \epsilon_k \sim \mathcal{N}(0, \sigma^{2}_0) \\
\end{cases} \\
& x_0 \sim \mathcal{N}(c_0, K_0)
\end{aligned}
\end{equation}

As we can see this is a random walk expressed by dynamic variable $x_k$ additionally submerged into white noise $\epsilon_k$. The covariance of this model as was mentioned in~\ref{sec:4_1} is: $\mathtt{Cov}[q_k^{(1)}(t_1), q_k^{(1)}(t_2)] = K_0 + q_0^2 \min(t_1,t_2)$. Now if we generalize this model to arbitrary time intervals it can be written as:

\begin{equation}\label{eq:21}
\begin{aligned}
&\begin{cases}
&\begin{bmatrix} x_k \\ \theta_k \end{bmatrix} = \begin{bmatrix}  1 & \Delta t_{k-1} \\ 0 & 1 \end{bmatrix}  
\begin{bmatrix} x_{k-1} \\ \theta_{k-1} \end{bmatrix} + \begin{bmatrix} q_k^{(1)} \\ q_k^{(2)} \end{bmatrix}\\
&y_k = \begin{bmatrix} 1 & 0 \end{bmatrix} \begin{bmatrix} x_k \\ \theta_k \end{bmatrix} + \epsilon_k, \quad \text{where:} \quad \epsilon_k \sim \mathcal{N}(0, \sigma^{2}_0)\\
\end{cases}\\
&\Delta t_{k-1} = t_k - t_{k-1}, \quad \text{it is assumed that $t_0 = 0$,} \\
&\begin{bmatrix} x_0 \\ \theta_0 \end{bmatrix} \sim \mathcal{N}\left( \begin{bmatrix} c_0 \\ \boxed{0} \end{bmatrix},
 \begin{bmatrix} K_0 & 0 \\ 0 & \boxed{0} \end{bmatrix} \right) \\
&\begin{bmatrix} q_k^{(1)} \\ q_k^{(2)} \end{bmatrix} \sim \mathcal{N}\left( \begin{bmatrix} 0 \\ 0 \end{bmatrix},
 \begin{bmatrix} q_0^2 \Delta t_{k-1} & 0 \\ 0 & \boxed{0} \end{bmatrix} \right) \\
\end{aligned}
\end{equation}

The parameters which are nullified with respect to the general model (LLLM) are denoted by boxes. We can see that the equation for $\theta_k$ is a redundant equation because $\theta_0$ is initialized as zero and corresponding noise term is also zero. The covariance function could also be obtained by using the formula for the general covariance function and putting to zeros the corresponding coefficients.

In the end of this section it is worth to mention that although the LLM is the simplest structural time series model, it can be successfully applied to the real world data~\cite[p. 16]{durbin_book}. 

\subsection{Local Linear Trend Model (LLLM)}

The next model we consider is called Local Linear Trend Model (LLLM). As was previously said it is the same as general model discussed in this section. The slope variable $\theta_k$ is changing by random walk, and the coordinate variable $x_k$ has also random walk components similarly to LLM. One can consider a simplification of LLLM: only $\theta_k$ changes by random walk but $x_k$ does not. As stated in~\cite[p. 44]{durbin_book} this simplified model produces smoother sample paths than general LLLM.

\section{Periodic and Quasi-Periodic (Cyclic) Modeling}\label{sec:5}

In the structural time series framework there are several models for periodicities and cycles (quasi-periodicities). We consider here the most popular model which is frequently used for cyclic modeling~\cite[p. 44]{durbin_book}: 

\begin{equation}\label{eq:22}
\begin{aligned}
&\begin{cases}
&\begin{bmatrix} x_k \\ x^{\star}_k \end{bmatrix} = \begin{bmatrix} \cos(\omega_c \Delta t_{k-1}) & \sin(\omega_c \Delta t_{k-1}) \\ -\sin(\omega_c \Delta t_{k-1}) & \cos(\omega_c \Delta t_{k-1}) \end{bmatrix}  
\begin{bmatrix} x_{k-1} \\ x^{\star}_{k-1} \end{bmatrix} + \\
&\qquad \qquad \qquad \qquad +\begin{bmatrix} q_k^{(1)} \\ q_k^{(2)} \end{bmatrix}\\
&y_k = \begin{bmatrix} 1 & 0 \end{bmatrix} \begin{bmatrix} x_k \\ x^{\star}_k \end{bmatrix} + \epsilon_k, \quad \text{where:} \quad \epsilon_k \sim \mathcal{N}(0, \sigma^{2}_0)\\
\end{cases}\\
&\Delta t_{k-1} = t_k - t_{k-1}, \quad \text{it is assumed that $t_0 = 0$,} \\
&\begin{bmatrix} x_0 \\ x^{\star}_0 \end{bmatrix} \sim \mathcal{N}\left( \begin{bmatrix} m_0 \\ m_0 \end{bmatrix},
 \begin{bmatrix} P_0 & 0 \\ 0 & P_0 \end{bmatrix} \right) \\
&\begin{bmatrix} q_k^{(1)} \\ q_k^{(2)} \end{bmatrix} \sim \mathcal{N}\left( \begin{bmatrix} 0 \\ 0 \end{bmatrix},
 \begin{bmatrix} g_0^2 \Delta t_{k-1} & 0 \\ 0 & g_0^2 \Delta t_{k-1} \end{bmatrix} \right) \\
\end{aligned}
\end{equation}

The presented equations are already a generalization of discrete time model which is usually encountered in the books~\cite{durbin_book}~\cite{harvey_book}, to the continuous time model. The $\Delta t_{i}$ are used to express the uneven time sampling. If the sampling is even all the $\Delta t_{i}$ equal to one.

Notice that the model is completely symmetric with respect to the vector $\mathbf{x} = [x_k, x_k^{\star}]$. The initial conditions are symmetric and the noise is symmetric. If we suppose no noise in the model then it is straightforward to show that the covariance function of $x_k$ is:

\begin{equation}\label{eq:23}
\mathit{Cov[x_k, x_{k+n}]} = P_0^2 \cos[\omega_c (t_{k+n} - t_{k})] 
\end{equation}

So, it is a periodic covariance function. The process $x_k$ can be considered as a random process where randomness originates only from the initial conditions. This process is also wide sense stationary since the covariance function depend on the difference of the time points. Again if we suppose that the the noise vector is absent from the dynamic model (i.e. $q_0^2 = 0$) then the $x_n$ variable is just a cosine wave. This can be deduced by considering $x_1$ which is a sum of cosine and sine with coefficients which are initial values: $x_0, x_0^{\star}$. This sum can be represented as a cosine wave where the phase depend on those coefficients. Also, we need to consider the property~(\ref{eq:24}) which is discussed soon. Hence, without extra white noise the $x_n$ is a cosine wave, however with the presence of white noise the deviations from the strict periodicity are possible.

\subsection{Quasi-Periodic (Cyclic) Covariance function}
Let's consider the dynamic matrix. Its spectral decomposition is written below:

\begin{equation}\label{eq:24}
\begin{aligned}
&A[\Delta t_{k-1}] = \\
&\begin{bmatrix} \cos(\omega_c \Delta t_{k-1}) & \sin(\omega_c \Delta t_{k-1}) \\ -\sin(\omega_c \Delta t_{k-1}) & \cos(\omega_c \Delta t_{k-1}) \end{bmatrix} = \\
&=\frac{1}{2} \begin{bmatrix}  1 & 1 \\ i & -i \end{bmatrix} \begin{bmatrix} e^{i \omega_c \Delta t_{k-1} } & 0 \\ 0 & e^{-i \omega_c \Delta t_{k-1}} \end{bmatrix} \begin{bmatrix}  1 & -i \\ 1 & i \end{bmatrix}
\end{aligned} 
\end{equation}

Using this it is easy to show that the property~(\ref{eq:9}) is valid again. Therefore, we conclude that all the results which are derived in the section~\ref{sec:4} and which are based on the property~(\ref{eq:9}) are also valid.  
In particular expressions~(\ref{eq:14}) and~(\ref{eq:15}) are valid which already give us the results for the covariance matrices of $\mathbf{z}_k = [ x_k, x^{\star}_k ]^T$. Repeating the same steps as are done to dive the covariance formula~(\ref{eq:17}) we can derive the similar formula for the cyclic model~(\ref{eq:22}). The 
derived covariance function consist of two parts as in Eq.~(\ref{eq:17}), however we must exclude the first row and the first column form the covariance matrix provided below similarly to formula~(\ref{eq:19}).
The two parts $\mathtt{Cov}_1$ and $\mathtt{Cov}_2$ are written below:   

\begin{equation}\label{eq:25}
\mathtt{Cov}_1[ \cdot ] = \mathcal{L}\{  \mathit{Cos} \{T\} \} \; D \; ( \, \mathcal{L}\{  \mathit{Cos} \{T\} \} \, )^T \\
\end{equation}

In this expression matrices $T$ and $D$ are exactly the same as in Eq.~(\ref{eq:19}). There are two new matrix operations which are nested: $\mathcal{L}\{ \cdot \}$ and $\mathit{Cos}\{ \cdot \}$. The first one leaves the lower triangular part (including the main diagonal) of the argument matrix intact, and put zeros to the upper-triangular part. The second one applies $\cos$ function element-wise to the matrix. 

Similarly,
\begin{equation}\label{eq:26}
\mathtt{Cov}_2[ \cdot ] = \mathcal{L}\{  \mathit{Sin} \{T\} \} \; D \; ( \, \mathcal{L}\{  \mathit{Sin} \{T\} \} \, )^T \\
\end{equation}

Here, $\mathit{Sin}\{ \cdot \}$ is used instead of $\mathit{Cos}\{ \cdot \}$ with the similar meaning - element-wise application of $\sin$ function to the argument matrix.

\begin{figure}[h]
\centering
\subfloat[~\mbox{Zero Noise:} \hfill ${\omega_c = 3, P_0=1, \underline{g^2_0=0}}$]{\includegraphics[width=0.44\columnwidth]{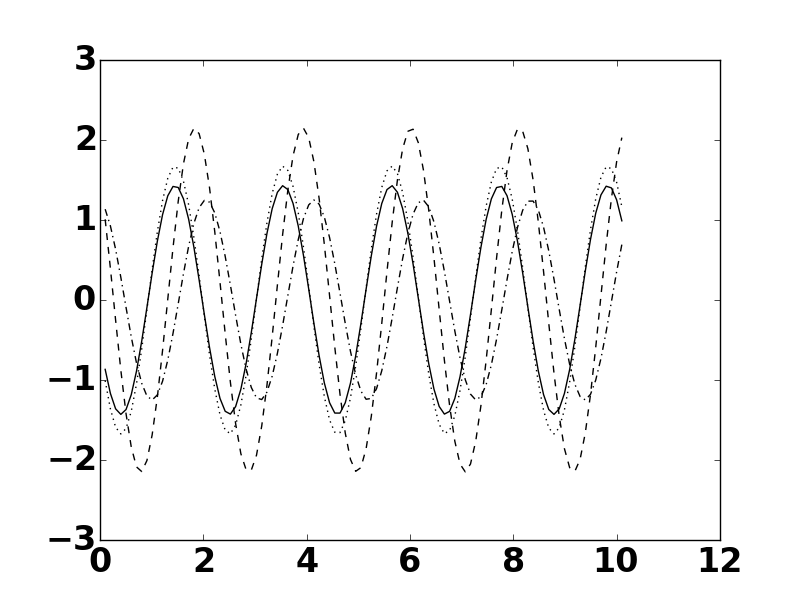}%
\label{fig_2a}}
\hfil
\subfloat[\mbox{Non-Zero Noise:} \hfill ${\omega_c = 3, P_0=1, g^2_0=1}$]{\includegraphics[width=0.44\columnwidth]{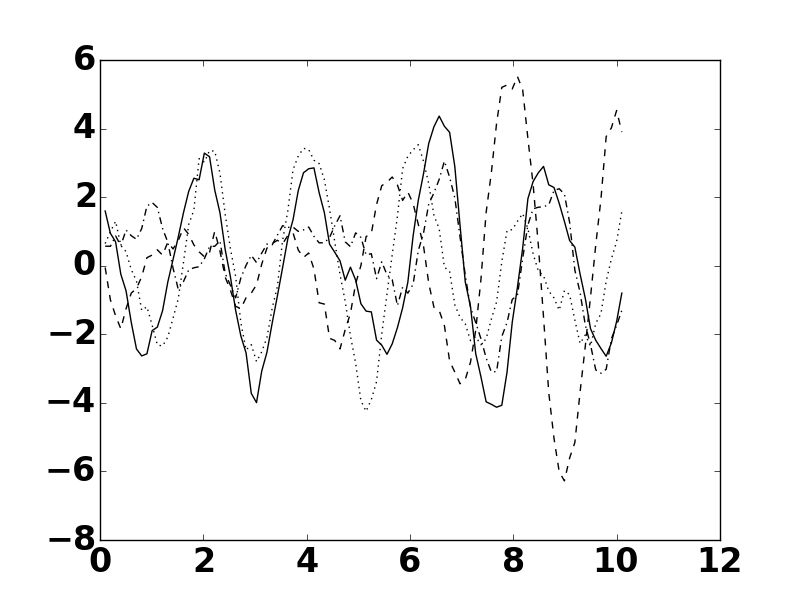}%
\label{fig_2b}}

\subfloat[Standard Periodic Covariance~\cite{rasmussen_2005}:]{\includegraphics[width=0.44\columnwidth]{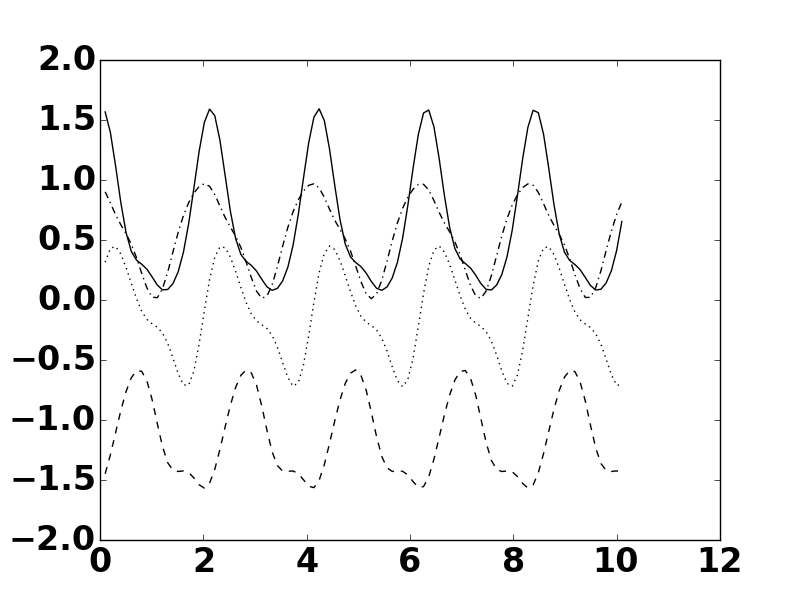}%
\label{fig_2c}}
\hfil
\subfloat[Quasi-Periodic Covariance~\cite{rasmussen_2005}:]{\includegraphics[width=0.44\columnwidth]{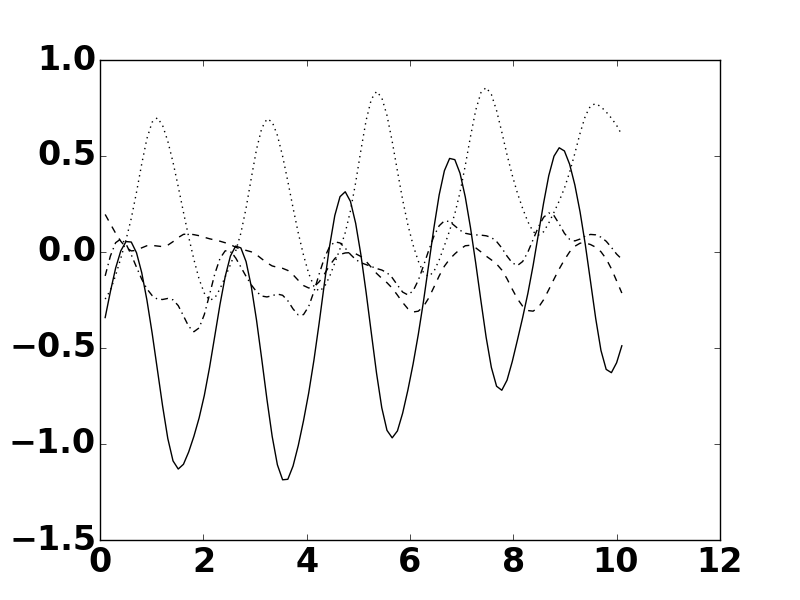}%
\label{fig_2d}}
\caption{Quasi-periodic (cyclic) sample paths. }
\label{fig_cyclic_paths}
\end{figure}

Thus, we obtained the expression for the covariance matrix of the quasi-periodic model Eq.~(\ref{eq:22}). Hence, it is now possible to model this cyclic state-space model as a Gaussian Process with the obtained covariance function. The GP sample paths with cyclic covariance function are shown on the Fig.~\ref{fig_2a}~\ref{fig_2b}.

If the data contains several frequencies or periodicities then the corresponding state-space models can be combined in the measurement equations for $y_k$, see subsection~\ref{sec:combine}. In GP regression this is equivalent to the summation of covariance functions.

Also, if the periodic pattern in the data is not close to cosine wave then we need to take more harmonics to model this pattern. Then we need to combine several frequencies: $\omega_c, 2\omega_c, \cdots, k\omega_c$ ($k$ harmonics) as described in the previous paragraph.

\subsection{Gaussian Periodic covariance function}

It is interesting to compare the periodicity modeling approach proposed above with the approach used in Gaussian Process Regression (GPR). In GPR there exist a periodic covariance function~\cite[p. 92]{rasmussen_2005} which is expressed as:

\begin{equation}\label{eq:27}
\mathit{Cov[t_1, t_2]} = \exp \left( - \sin^2 \left( \frac{\omega_c (t_1 - t_2)}{ 2} \right) \right)
\end{equation}

This is also a periodic covariance function with a frequency $\omega_c$. Sample paths from GP with periodic covariance are presented on Fig.~\ref{fig_2c}. Since the covariance function is periodic it is possible to represent it as a Fourier series with a harmonics $\omega_c, 2\omega_c, 3\omega_c, \cdots $. This is exactly the case which can be represented by combining state-space models and which is described in the previous subsection. Thus, the periodic covariance function used in GPR can be represented by equivalent random process in the state-space form. It is done in the paper~\cite{solin_2014}. 

In the same paper the question of representing the quasi-periodic covariance function is also discussed. The quasi-periodic covariance function is a multiplication of some stationary covariance functions (e.g. Matern covariance)~\cite{solin_2014} and the periodic one in Eq.~(\ref{eq:27}). The random process which is modeled by quasi-periodic covariance function has no fixed period, the period length is fluctuating. Sample paths of quasi-periodic covariance are shown on Fig.~\ref{fig_2d}. By using noise $\mathbf{q_k}$ we also deviate from strict periodicity, however there is no direct correspondence between model in Eq.~(\ref{eq:22}) and quasi-periodic covariance function in the paper~\cite{solin_2014}. This question requires further investigation and is not touched here anymore.

\section{Damped Trend Model}\label{sec:6}

In this section we consider damping trend model. It is similar to the general model Eq.~(\ref{eq:7}), except that a slope gradually decreases. Here we present only the dynamic equation for this model because the rest is the same as in Eq.~(\ref{eq:7}).

\begin{equation}\label{eq:28}
\begin{bmatrix} x_k \\ \theta_k \end{bmatrix} = \begin{bmatrix}  1 & \Delta t_{k-1} \\ 0 & \boxed{\phi} \end{bmatrix}  
\begin{bmatrix} x_{k-1} \\ \theta_{k-1} \end{bmatrix} + \begin{bmatrix} q_k^{(1)} \\ q_k^{(2)} \end{bmatrix}
\end{equation}

The damping factor is denoted by the box around it. It must satisfy $0 < \phi < 1$ so that the trend to be damping. 

Next we present the covariance function of this damping trend. The derivation is omitted because it is very similar to the derivation LLLM model Eq.~(\ref{eq:17}). The first part of the covariance $\mathtt{Cov}_1[ \cdot ]$ is the same as in Eq.~(\ref{eq:18}). The second is also similar to Eq.~(\ref{eq:19}) except that matrix $T$ must be substituted to:

\begin{equation}\label{eq:29}
\begin{aligned} 
&T = \\
&\begin{bmatrix}
0 & 0 & 0 & 0\\
1 & 0 & 0 & 0\\
1 & \phi & 0 & 0\\
1 & \phi & \phi^2 & 0\\  
\end{bmatrix}
\begin{bmatrix}
0 & 0 & 0 & 0\\
\Delta t_{0} & 0 & 0 & 0\\
\Delta t_{1}& \Delta t_{1} & 0 & 0\\
\Delta t_{2}& \Delta t_{2} & \Delta t_{2} & 0\\  
\end{bmatrix} 
\begin{bmatrix}
0 & 0 & 0 & 0\\
1 & 0 & 0 & 0\\
0 & \frac{1}{\phi} & 0 & 0\\
0 & 0 & \frac{1}{\phi^2} & 0\\  
\end{bmatrix}\\
\end{aligned}
\end{equation}

\begin{figure}[h]
\centering
\subfloat[~\mbox{No noise: $\phi=0.95$ } \hfill ${K_0 = P_0=1,\; \underline{q^2_0=g^2_0=0}}$]{\includegraphics[width=0.44\columnwidth]{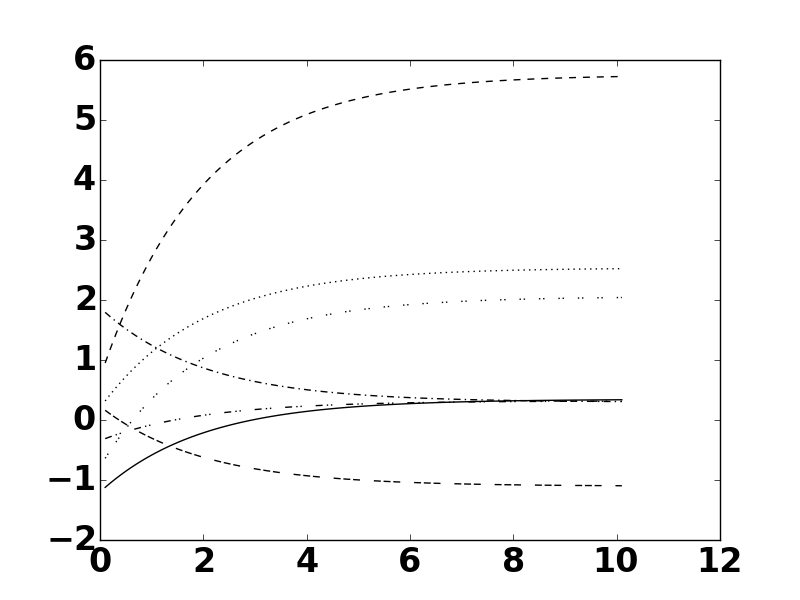}%
\label{fig_3a}}
\hfil
\subfloat[\mbox{Non-Zero Noise: $\phi=0.95$} \hfill ${K_0 = P_0=1, \; q^2_0=g^2_0=0.05}$]{\includegraphics[width=0.44\columnwidth]{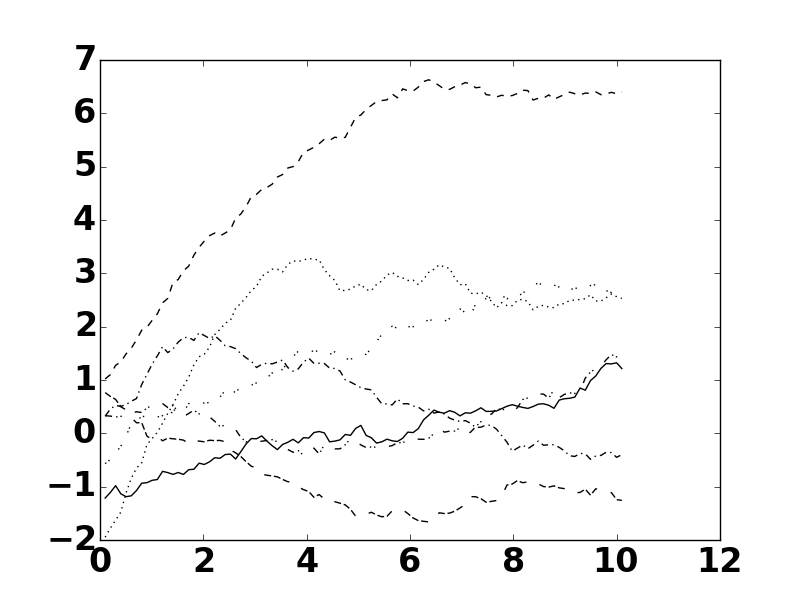}%
\label{fig_3b}}

\caption{Damped sample paths.}
\label{fig_damped_paths}
\end{figure}

As before, to obtain the final covariance we must discard the first row and the first column from the resulting covariance. It is worth noting that the model is not completely adapted to the continuous time. The reason is that damping factor $\phi$ does not depend on the time interval $\Delta t_{k-1}$ between two consecutive measurements $y_k$. So, strictly speaking the covariance Eq.~(\ref{eq:29}) is valid only when all $\Delta t_i$ are the same. It is possible to extend the derived covariance to cover the general case as well, however for simplicity of presentation and space constrains it is not done here. Sample paths from GP with a damped trend covariance are given on Fig.~\ref{fig_damped_paths}.

\section{External Variables}
So far we have considered the modeling of $y_k$ with respect to time. These might include local level model, deterministic or stochastic trend, one or more periodicities etc.. These time patterns are modeled by state-space model for variable $x_k$. Quite often, there might be other explanatory variables e.g. day of the week. We can also include them into the model. Suppose that $y_k$ depends linearly on a set of explanatory variables $\mathbf{z} = [z_1, z_2, \cdots z_m]^T$:

\begin{equation}\label{eq:30} 
y_k = x_k + \mathbf{b}^T \mathbf{z}_k + \epsilon_k
\end{equation}

In the formula above $\mathbf{b}$ is some vector of parameters. Denote also that $f_k = \mathbf{b}^T \mathbf{z}_k$. If we assume that the vector $\mathbf{b}$ is a vector of constant but unknown parameters we again can express this model both in state-space and in GP forms.
To express in the state-space form it is enough to assign $\mathbf{b}$ as a state variable with unit dynamic (transition) matrix and no noise. Then we need to combine this state-space model with the one for $x_k$. It is shown in the Sec.~\ref{sec:combine} how to do that. 

It is easy to check that if $x_k$ and $f_k$ are independent random processes:
\begin{equation}\label{eq:31}
\mathit{Cov[y_k, y_{k+n}]}  = \mathit{Cov[x_k, x_{k+n}]} + \mathit{Cov[f_k, f_{k+n}]}
\end{equation}
So, the covariance function is the sum of two covariance functions for $x_k$ and $z_k$. In our case, $x_k$ and $\mathbf{b}^T \mathbf{z}_k$ are independent. The randomness to the second process is introduced only through the prior distribution of parameters $\mathbf{b}$, which is independent of a randomness in $x_k$. We can also see that the dependency of $\mathbf{z}_k$ is exactly Bayesian Linear Regression introduced in the Sec.~\ref{sec:blr}. The only difference is that now $\mathbf{z}_k$ is possibly multidimensional vector. Anyway, the covariance function of BLR part is:

\begin{equation}\label{eq:32}
\mathbf{f} \sim \mathcal{GP} \left( 0, z_0^2 Z Z^T  \right)
\end{equation}
where it is assumed that:
\begin{equation}
\mathbf{b} \sim \mathcal{N}(0,z_0^2 I) \quad -\text{prior}
\end{equation}
And $Z$ is a matrix composed of vectors $\mathbf{z}_k$ row-wise. This is analogous to the Eq.~(\ref{eq:5}) except that the noise term is missing in this covariance.s

\section{ARMA Models}

The discrete WSS random processes are frequently modeled as an Auto-Regressive Moving-Average (ARMA) process~\cite{box_book_2008}:

\begin{equation}\label{eq:33}
x_n + a_1 x_{n-1} + a_2 x_{n-2} + \cdots + a_p x_{n-p} = b_0 \xi_0 + b_1 \xi_1 + \cdots + b_q \xi_q
\end{equation}

Where $a_i$ and $b_i$ are some real valued coefficients, $\xi_i$ are independent Gaussian white noise with unit variance. It is straightforward to write this ARMA(p,q) model in the state-space form~\cite{durbin_book}. We do not present it here due to the space constraints.

The process in Eq.~(\ref{eq:33}) is stationary under some conditions on the coefficients and its power spectrum
is:

\begin{equation}\label{eq:34}
P_x (\omega) = \frac{| B_q(e^{i\omega})|^2}{| A_p(e^{i\omega}) |^2} 
\end{equation}
where $A_p(e^{i\omega})$ and $B_q(e^{i\omega})$ are polynomials with corresponding coefficients from Eq.~(\ref{eq:33}). For instance:
\begin{equation}\label{eq:35}
A_p(e^{i\omega}) = 1 + a_1 e^{i\omega} + a_2 e^{i2\omega} + \cdots + a_p e^{ip\omega}
\end{equation}
Also we can see that the this power spectrum is periodic with the period $2\pi$ because exponent is the periodic function with this period. 

The ARMA processes can be generalized to the continuous time: ARMA process in continuous time has rational spectral density of the form Eq.~(\ref{eq:34}) except that instead of the argument $e^{i\omega}$ the argument $i\omega$ must be used. There must be extra requirements in order that the rational function represents a power spectrum of a random process. Namely, numerator and denominator do not have common roots, there can not be real roots, and that $p \geq q + 1$~\cite[p. 133]{yaglom_book1987}. 

It is possible to write a covariance function of this random process by computing the Fourier transform of the power spectrum. The covariance function is a sum of the terms which depend on the roots of the denominator of the $|A_p(i\omega)|^2$. Since roots appear in conjugate pairs and every pair must be taken into account only once we consider only the roots with positive imaginary parts. We write here formulas for the case when all the roots are of unit multiplicity, for the general case see e.g.~\cite{yaglom_book1987}. Each fully imaginary root (with positive imaginary part) $i \alpha_k$ brings the following term to the sum:

\begin{equation}\label{eq:36}
r_1(\tau) = C e^{-\alpha_k |\tau|}
\end{equation}
This is an exponential covariance function. Each complex root (with positive imaginary part) $\alpha_k^{(1)} + i \alpha_k^{(2)}$ introduces a term:

\begin{equation}\label{eq:37}
\begin{aligned}
&r_2(\tau) = C e^{-\alpha_k^{(2)} |\tau|} cos(\alpha_k^{(1)}|\tau| - \psi) \qquad \text{where:} \\
&|\psi| \leq \tan^{-1}(\alpha_k^{(2)} / \alpha_k^{(1)}) \quad -\text{some phase shift.}\\
\end{aligned}
\end{equation}

For instance the last covariance function is a correlation function of continuous ARMA(2,1) process. More information on this topic can be found in~\cite{yaglom_book1987},~\cite{ihara_book_1993},~\cite{rasmussen_2005}.

\section{Experiments}

In this section we perform a number of basic experiments in order to demonstrate that the derived in Sec.~\ref{sec:4},~\ref{sec:5},~\ref{sec:6} covariance functions are applicable in the GP regression framework and to show that the results are equivalent to state-space modeling. The proposed kernels are applied to several artificially generated datasets and it is shown that GP regression results are meaningful. Furthermore, we compare the GP regression approach and the state-space approach for the Nile Water Level~\cite{cobb_78} dataset which is frequently used in the time series literature. It is shown on the simple example of LLM model from~\ref{sec:4} that the modeling results are equivalent.

All new kernels proposed in this paper have been implemented as an add-ons to the \textit{GPy toolbox}. This a powerful toolbox for Gaussian Process modeling and inference~\cite{gpy_2014}. Crucial part of GP inference is finding hyper-parameters of a kernel. A standard way to do this is to find maximum (MAP estimate) of marginal log-likelihood~\cite[p. 112]{rasmussen_2005}. In the subsequent experiments maximum is searched by BFGS algorithm. Since marginal log-likelihood is non-convex function, each optimization procedure is run 10 times with different random initial conditions. The hyper-parameters which produce the highest marginal log-likelihood are considered as final answer.  

\begin{figure}[h]
\centering
\subfloat[~\mbox{Dataset 1: $q_0^2 = g_0^2 = 0$ } ]{\includegraphics[width=0.44\columnwidth]{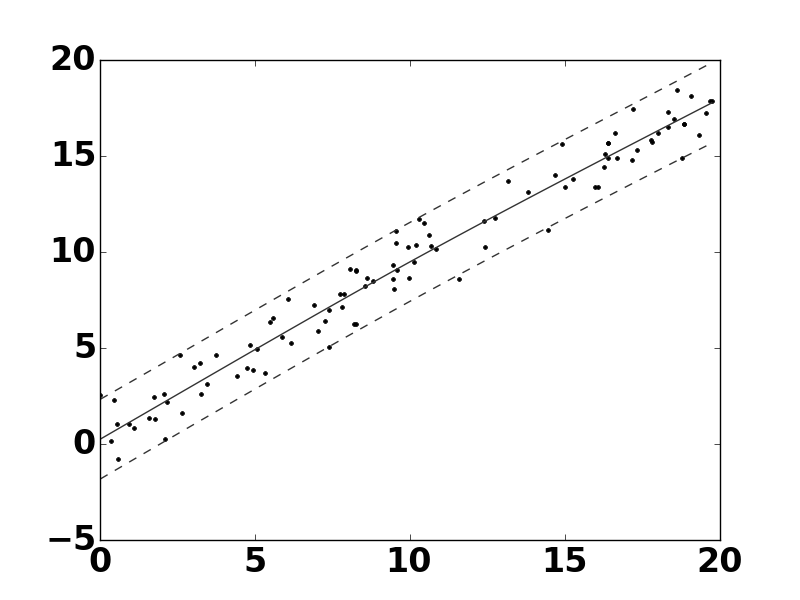}%
\label{fig_4a}}
\hfil
\subfloat[\mbox{Dataset 2: $q_0^2 = g_0^2 = 1$}]{\includegraphics[width=0.44\columnwidth]{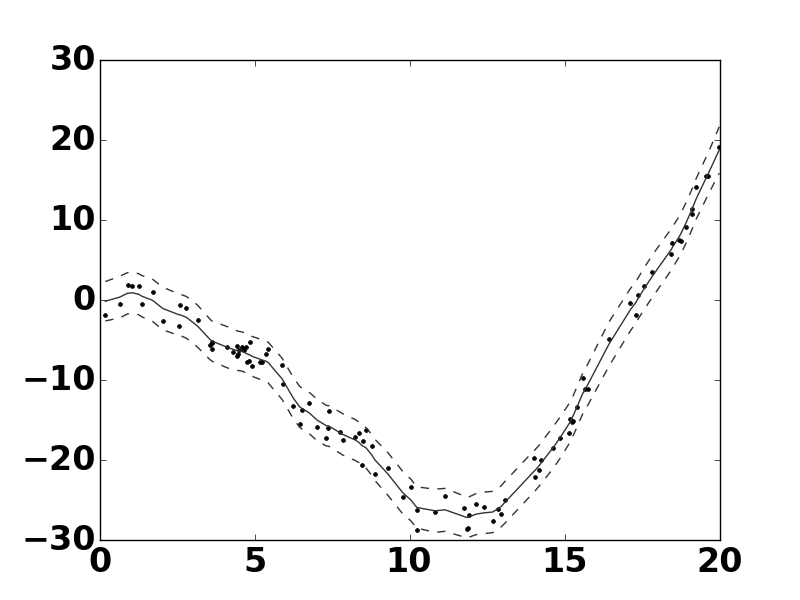}%
\label{fig_4b}}

\caption{GP regression with general state-space kernel Eq.~(\ref{eq:17})}
\label{fig_gen_ss_regress}
\end{figure}

\begin{table}[h]
\renewcommand{\arraystretch}{1.3}
\caption{GP regression with general state-space kernel Eq.~(\ref{eq:11})}
\label{tab:1}
\centering
\begin{tabular}{|p{0.6cm}|p{0.9cm}|p{1.4cm}||p{0.9cm}|p{1.4cm}|}
\hline
 & \multicolumn{2}{|c||}{Dataset 1} & \multicolumn{2}{c|}{Dataset 2}\\
\hline
Param. name & True value & MAP estimation & True value & MAP estimation\\
\hline
$K_0$ & 1.0 & 0.14 & 1.0 & $5.79*10^{-7}$\\
$P_0$ & 1.0 & 0.09 & 1.0 & $2.02*10^{-7}$ \\
$q_0^2$ & 0.0 & $6.05*10^{-8}$ & 1.0 & 1.53\\
$g_0^2$ & 0.0 & $8.08*10^{-4}$ & 1.0 & 3.17\\
$\sigma_0^2$ & 1.0 & 1.08 & 1.0 & 1.39\\
\hline
\end{tabular}
\end{table}

\begin{figure}[h]
\centering
\subfloat[~\mbox{Dataset 1: $g_0^2 = 0$ } ]{\includegraphics[width=0.44\columnwidth]{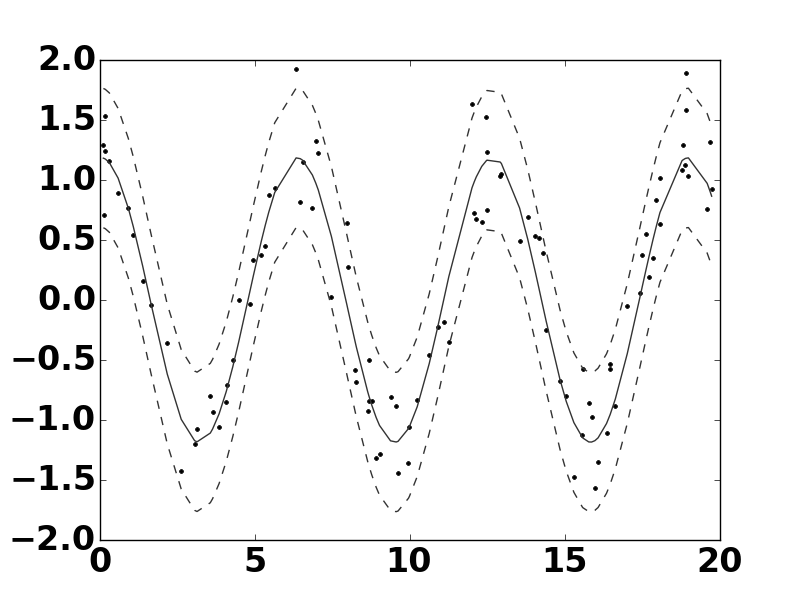}%
\label{fig_5a}}
\hfil
\subfloat[\mbox{Dataset 2: $g_0^2 = 1$}]{\includegraphics[width=0.44\columnwidth]{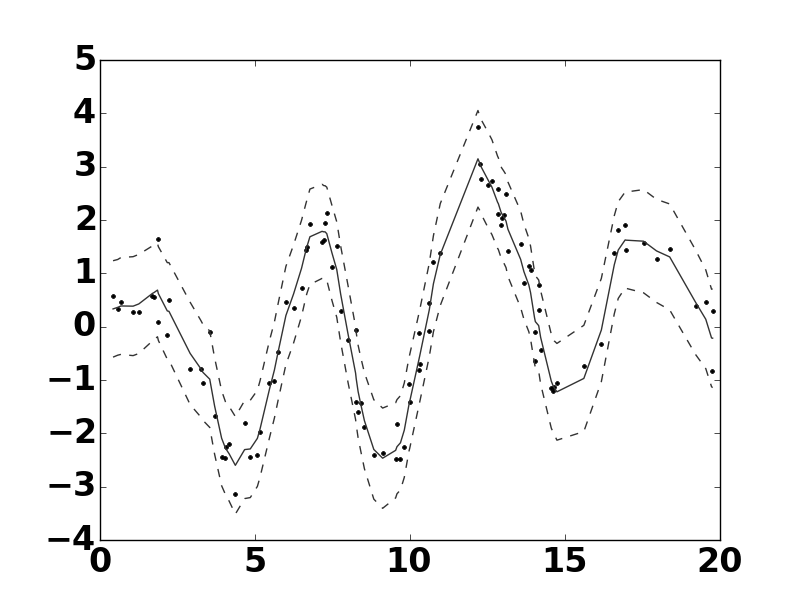}%
\label{fig_5b}}

\caption{GP regression with quasi-periodic kernel Eq.~(\ref{eq:25}),~(\ref{eq:26})}
\label{fig_cyclic_regress}
\end{figure}

\begin{table}[h]
\renewcommand{\arraystretch}{1.3}
\caption{GP regression with periodic kernel Eq.~(\ref{eq:19}),~(\ref{eq:20})}
\label{tab:2}
\centering
\begin{tabular}{|p{0.6cm}|p{0.9cm}|p{1.6cm}||p{0.9cm}|p{1.4cm}|}
\hline
 & \multicolumn{2}{|c||}{Dataset 1} & \multicolumn{2}{c|}{Dataset 2}\\
\hline
Param. name & True value & MAP estimation & True value & MAP estimation\\
\hline
$\omega_c$ & 1.0 & 0.99 & 1.0 & 1.04 \\
$P_0$ & 1.0 & 0.70 & 1.0 & $1.64*10^{-7}$ \\
$g_0^2$ & 0.0 & $1.23*10^{-15}$ & 0.1 & 0.44\\
$\sigma_0^2$ & 0.1 & 0.09 & 0.1 & 0.16\\
\hline
\end{tabular}
\end{table}

\begin{figure}[h]
\centering
\subfloat[~\mbox{Dataset 1: $q_0^2 = g_0^2 = 0$ } ]{\includegraphics[width=0.44\columnwidth]{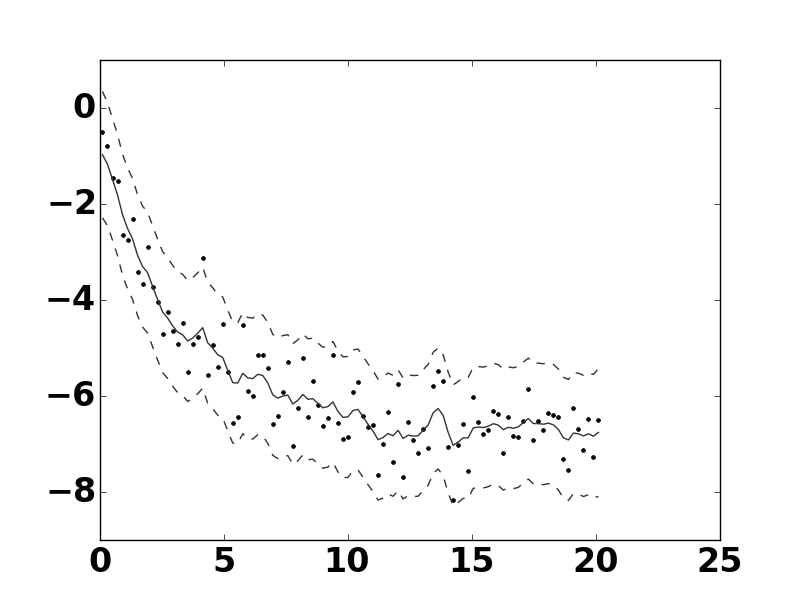}%
\label{fig_6a}}
\hfil
\subfloat[\mbox{Dataset 2: $q_0^2 = g_0^2 = 0.01$}]{\includegraphics[width=0.44\columnwidth]{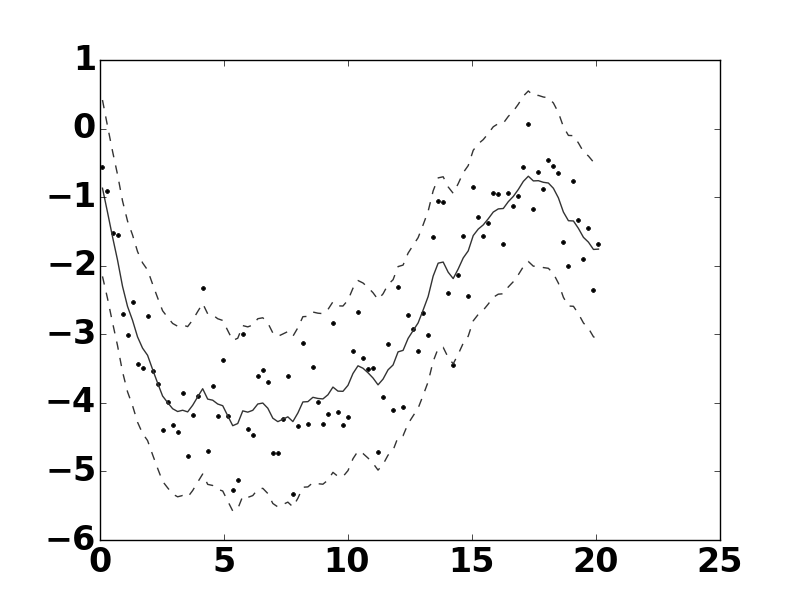}%
\label{fig_6b}}

\caption{GP regression with damped trend kernel Eq.~(\ref{eq:29})}
\label{fig_damped_regress}
\end{figure}

\begin{table}[h]
\renewcommand{\arraystretch}{1.3}
\caption{GP regression with damped trend kernel Eq.~(\ref{eq:29})}
\label{tab:3}
\centering
\begin{tabular}{|p{0.6cm}|p{0.9cm}|p{1.4cm}||p{0.9cm}|p{1.4cm}|}
\hline
 & \multicolumn{2}{|c||}{Dataset 1} & \multicolumn{2}{c|}{Dataset 2}\\
\hline
Param. name & True value & MAP estimation & True value & MAP estimation\\
\hline
$\phi$ & 0.94 & 0.02 & 0.94 & 0.68\\
$K_0$ & 3.0 & 1.50 & 3.0 & 0.4\\
$P_0$ & 1.0 & 0.5 & 1.0 & 0.85 \\
$q_0^2$ & 0.0 & 0.46 & 0.01 & 0.3\\
$g_0^2$ & 0.0 & 0.5 & 0.01 & 0.19\\
$\sigma_0^2$ & 0.4 & 0.33 & 0.4 & 0.33\\
\hline
\end{tabular}
\end{table}

The first experiment is designed to test the general state-space covariance Eq.~(\ref{eq:17}). Two datasets from the model Eq.~(\ref{eq:7}) are generated each containing 100 points. In the first dataset the parameters $q_0^2 = 0, g_0^2 = 0$ which means the absence of noise of the dynamic model and equivalence to BLR Eq.~(\ref{eq:4}).  In the second dataset noise parameters are $q_0^2 = 1, g_0^2 = 1$, so they are non-zero. All the remaining parameters $K_0, P_0, \sigma_0^2$ equal to 1, and $c_0, m_0$ equal to zero. The results of GP regression modeling with general state-space covariance Eq.~(\ref{eq:17}) are presented in Table~\ref{tab:1} and Figure~\ref{fig_gen_ss_regress}. 

As we can see the Table~\ref{tab:1} and Figure~\ref{fig_gen_ss_regress}. The modeling provides quite feasible results. All parameters except $K_0$ and $P_0$ are estimated with reasonable accuracy for this kind of modeling.
The large error in estimation of $K_0$ and $P_0$ probably stems from the fact that the values of corresponding random variables are observed only once during the generation of initial state variables. This situation is quite typical for subsequent experiments as well. 

Similar experiment is performed for the periodic (or cyclic) covariance function which is a sum of Eq.~(\ref{eq:25}) and Eq.~(\ref{eq:26}). Dataset 1 which is generated with no noise in dynamic model correspond to purely periodic random process. The dataset 2 which has this noise correspond to quasi-periodic or cyclic behavior.
Results of GP regression with periodic kernel is presented in Table~\ref{tab:2} and Figure~\ref{fig_cyclic_regress}. They are also very reasonable. It is more important that noise levels and angular frequency of oscillations are estimated well.

The last kernel we experimented with is the damped trend model in Eq.~(\ref{eq:29}). If there is no noise in the
dynamic equation then the data generated by the model in Eq.~(\ref{eq:28}) is the damped trend, if noise is present then the generated data is more complex. Experimental results for this two cases are present in Table~\ref{tab:3} and Figure~\ref{fig_damped_regress}. We can see that the estimated parameters are much less accurate. Perhaps, this happens because this is the most complex model we have considered so far (in terms of number of parameters) and the same data can be generated by several different sets of parameters. Therefore, the true set is harder to identify. Anyway the plots on Figure~\ref{fig_damped_regress} show quite feasible results.

\begin{figure}[h]
\centering
\subfloat[~\mbox{ State-space model } ]{\includegraphics[width=0.44\columnwidth]{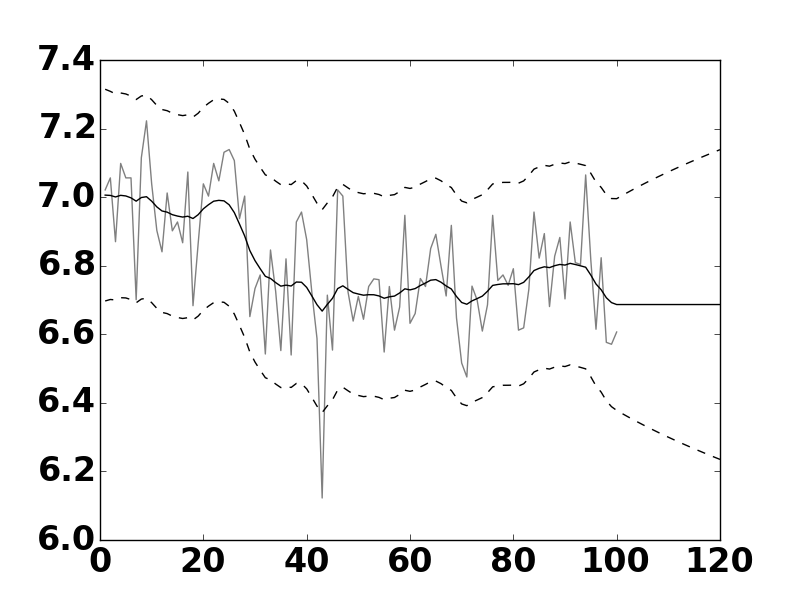}%
\label{fig_7a}}
\hfil
\subfloat[\mbox{GP regression}]{\includegraphics[width=0.44\columnwidth]{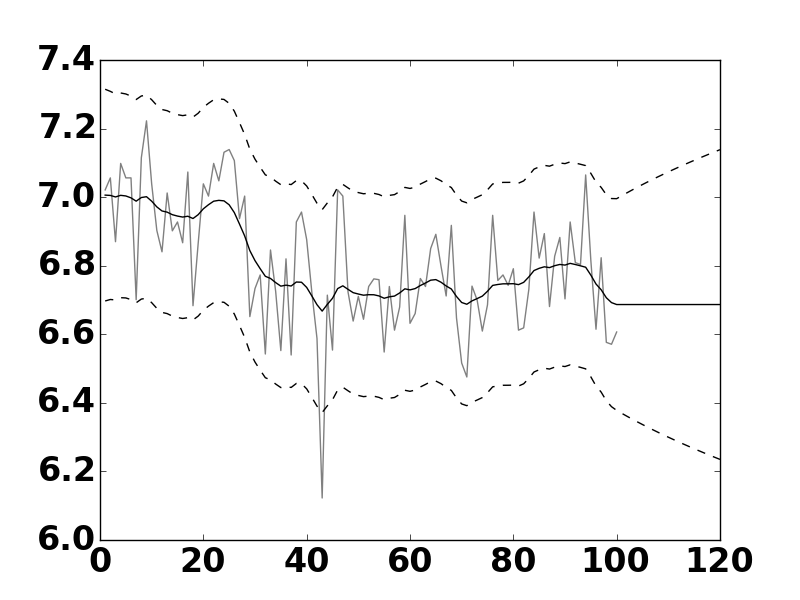}%
\label{fig_7b}}

\caption{Comparison of time series forecasting of GP regression and state-space model.}
\label{fig_last_regress}
\end{figure}

Finally, we want to demonstrate that the GP regression approach complemented with the kernels proposed in this paper is completely equivalent with state-space modeling approach. This is demonstrated on the classical Nile Water Level dataset~\cite{cobb_78} which contains 100 years (100 data points) of measurements. The state-space inference is performed by Kalman Filtering (KF) and Rauch-Tung-Striebel (RTS) smoother. We have taken the simple Local Level Model (LLM) from Eq.~(\ref{eq:21}). Often this model is used as a starting point for time series analysis. The results of modeling and forecasting of Nile dataset are presented on Figure~\ref{fig_last_regress}. From the figure it is impossible to see any difference between approaches. Analysis of numerical data, which is not presented here also shows that the difference is negligible. Hence, we have shown experimentally for one model that state-space approach and GP regression can be used interchangeably depending on the modeler's preferences and other relevant considerations.

\section{Conclusion}
In this paper we have considered the question of transforming popular state-space models (or structural time series models) into corresponding Gaussian Processes. The reverse transformation is studied in e.g.~\cite{solin_2014} and references there in. We have considered general Local Linear Trend Model (LLLM) and its simplifications, quasi-periodic (cyclic) state-space model, damped trend model. At first, these models are written in the continuous time forms and then corresponding GP kernels are derived. Other widely used models like ARMA, external variables and model combinations have been mentioned and the way to construct GP kernels for them have been shown. 

We have demonstrated the correctness and feasibility of the GP regression with novel kernels on the several synthetic datasets and equivalence with state-space modeling is shown on a real world dataset.

Thus, this paper makes a bridge between state-space and GP modeling and forecasting of time series data. It allows experts in either of the fields to look at their models from the other point of view and share the ideas between those approaches of modeling.

\bibliographystyle{IEEEtran}
\bibliography{IEEEabrv,ss_bibliography}
\end{document}